\documentclass[10pt,twocolumn,letterpaper]{article}

\usepackage{cvpr}
\usepackage{times}
\usepackage{epsfig}
\usepackage{graphicx}
\usepackage{amsmath}
\usepackage{amssymb}
\usepackage{color}
\usepackage{multirow}

\usepackage[pagebackref=true,breaklinks=true,letterpaper=true,colorlinks,bookmarks=false]{hyperref}

\cvprfinalcopy 

\def\cvprPaperID{31} 

\ifcvprfinal\fi

\begin{document}

\title{Pedestrian Detection in Thermal Images using Saliency Maps}

\newcommand*\samethanks[1][\value{footnote}]{\footnotemark[#1]}

\author{Debasmita Ghose\thanks{Authors contributed equally}\qquad Shasvat M. Desai\samethanks\qquad Sneha Bhattacharya\samethanks\qquad Deep Chakraborty\samethanks \\ Madalina Fiterau\qquad Tauhidur Rahman\\
College of Information and Computer Sciences, University of Massachusetts, Amherst, MA 01002\\
{\tt\small \{dghose, shasvatmukes, snehabhattac, dchakraborty, mfiterau, trahman\}@cs.umass.edu}
}

\maketitle
\thispagestyle{empty}

\begin{abstract}
Thermal images are mainly used to detect the presence of people at night or in bad lighting conditions, but perform poorly at daytime. To solve this problem, most state-of-the-art techniques employ a fusion network that uses features from paired thermal and color images. Instead, we propose to augment thermal images with their saliency maps, to serve as an attention mechanism for the pedestrian detector especially during daytime. We investigate how such an approach results in improved performance for pedestrian detection using only thermal images, eliminating the need for paired color images. For our experiments, we train the Faster R-CNN for pedestrian detection and report the added effect of saliency maps generated using static and deep methods (PiCA-Net and $R^3$-Net). Our best performing model results in an absolute reduction of miss rate by 13.4\% and 19.4\% over the baseline in day and night images respectively. We also annotate and release pixel level masks of pedestrians on a subset of the KAIST Multispectral Pedestrian Detection dataset, which is a first publicly available dataset for salient pedestrian detection.
\end{abstract}

\section{Introduction}
\label{sec:intro}

Detecting the presence and location of pedestrians in a scene is a crucial task for several applications such as video surveillance systems \cite{wang2014scene} and autonomous driving \cite{geigerwe}. Despite the challenges associated with it, such as low resolution and occlusion, pedestrian detection has already been successfully studied widely in color images and videos using state-of-the-art deep learning techniques for object detection and/or semantic segmentation \cite{cai2015learning, li2018scale, du2017fused, SDS-RCNN}. Color images of reasonable quality are good for detecting pedestrians during the day. Thermal images, however, are very useful in detecting pedestrians in conditions where color images fail, such as nighttime or under bad lighting conditions. 
This is because at nighttime, thermal cameras capture humans distinctly as they are warmer than their surrounding objects. During the day however, there are other objects in the surroundings which are as warm as or warmer than humans, making them less distinguishable. Therefore, there appears to be a clear complementary potential between color and thermal images. In order to exploit this complementary potential, there has been a lot of work on building fusion architectures combining color and thermal images \cite{wagner2016multispectral, xu2017learning, liu2016multispectral, li2019illumination}. But color-thermal image pairs might not always be available, as they are expensive to collect and need image registration to be completely accurate. Misaligned imagery can also reduce the performance of a detector that leverages multiple data modalities. This motivates us to use only thermal images for the task of pedestrian detection. 

To address the challenge of pedestrian detection in thermal images, especially during daytime, we propose the use of saliency maps. Koch and Ulman~\cite{funda-saliency} define saliency at a given location by how different this location is from its surroundings in color, orientation, motion, and depth. Looking for salient objects in a scene can be interpreted as being a visual attention mechanism which illuminates pixels belonging to salient objects in a given scene. We therefore hypothesize that using saliency maps along with thermal images would help us improve the performance of state-of-the-art pedestrian detectors, especially on thermal images captured during the day. To test our hypothesis, we first establish a baseline by training a state-of-the-art object detector (Faster R-CNN \cite{ren2015faster}) to detect pedestrians solely from thermal images in the KAIST Multispectral Pedestrian dataset ~\cite{kaist_benchmarkpaper}. We then train pedestrian detectors on thermal images augmented with their saliency maps generated using static and deep learning techniques (PiCA-Net\cite{picanet} and $R^3$-Net\cite{r3net}).
Our experiments show that the pedestrian detector trained using this augmentation technique outperforms the baseline by a significant margin. Moreover, since deep saliency networks require pixel level annotations of salient objects, we annotate a subset of the KAIST multispectral pedestrian dataset \cite{kaist_benchmarkpaper} with pixel level masks for pedestrian instances to facilitate research on salient pedestrian detection. 

The key contributions of this paper are as follows: 
\begin{enumerate}
    \item To the best of our knowledge, this is the first paper to show the impact of saliency maps in improving the performance of pedestrian detection in thermal images.
    \item We release the first pixel level annotations for a multispectral pedestrian detection dataset and provide saliency detection benchmarks on it using state-of-the-art networks.
\end{enumerate}

The rest of the paper is organized as follows: Section~\ref{sec:related} reviews existing work on pedestrian detection in color and multispectral images and methods for saliency detection in images. Section~\ref{sec:approach} outlines the baseline method for pedestrian detection and our efforts to improve it using saliency maps. We also present a new salient pedestrian detection dataset that we annotated for this purpose. In Section~\ref{sec:exp} we report implementation details, benchmarks for our novel dataset and evaluate the performance of different techniques qualitatively and quantitatively. Finally, we present our conclusions and future work in Section~\ref{sec:conc}.

\section{Related Work}
\label{sec:related}
\textbf{Pedestrian detection.}
Traditionally, pedestrian detectors involved the use of hand crafted features and algorithms such as ICF \cite{icf}, ACF \cite{fast_feature_pyramid} and LDCF \cite{LDCF}. Deep learning approaches have however been more successful recently. Zhang \etal\cite{isfrnndoingwell} investigate the performance of the Faster R-CNN \cite{ren2015faster} for the task of pedestrian detection. Sermanet \etal\cite{82294fc46f3a4f75aa85a3f344b0358a} introduce the use of multistage unsupervised features and skip connections for pedestrian detection. Li \etal\cite{li2018scale} introduce Scale Aware Fast R-CNN which uses built-in sub-networks to detect pedestrians at different scales. In \cite{SDS-RCNN}, Brazil \etal introduce SDS R-CNN which uses joint supervision on pedestrian detection and semantic segmentation to illuminate pedestrians in the frame. This motivates us to use saliency maps as a stronger attention mechanism to illuminate pedestrians for detection. 

With the release of several multispectral datasets \cite{kaist_benchmarkpaper, osu_paper, zhang2015vais, wu2014thermal}, multimodal detectors have seen increasing popularity.
To exploit the complementary potential between thermal and RGB images, Liu \etal \cite{liu2016multispectral} introduce a fusion method based on the Faster R-CNN. Li \etal \cite{li2019illumination} introduce Illumination Aware Faster R-CNN which adaptively integrates color and thermal sub-networks, and fuses the results using a weighting scheme depending on the illumination condition. Region Re-construction Network is introduced in \cite{xu2017learning} which models the relation between RGB and thermal data using a CNN. These features are then fed into a Multi-Scale Detection Network, for robust pedestrian detection. In our approach however, we use solely the thermal images and not their color counterparts. 

\textbf{Saliency detection.}
Salient object detection aims to highlight the most conspicuous object in an image and a substantial number of methods have been developed for it over the past few decades. One of the earliest works on saliency detection was presented in \cite{funda-saliency}, inspired by the visual system of primates which shift focus to most conspicuous objects across the visual scene. Traditional saliency detection methods involved using methods like global contrast \cite{global_contrast_saliency}, local contrast \cite{Klein2011CentersurroundDO} and other hand crafted features like colour and texture \cite{learning_to_detect_salient, saliency_detection_graph_based}. Methods described in \cite{Hou07saliencydetection:, montabone2010human} form the basis for our experiments using static saliency. A complete survey of these methods is available in \cite{DBLP:huaizu}. 

Recent works typically use CNNs for salient object detection. DHSNet \cite{dhsnet} first learns global saliency cues such as global contrast, objectness, and compactness, and then uses a novel hierarchical convolutional neural network to refine the details of the saliency maps using local context information. 
The use of short connections to the skip layer structure of a Holistically-Nested Edge Detector is introduced in \cite{hou2017deeply}. Amulet~\cite{amulet} integrates multi-level features at multiple resolutions and learns to predict saliency maps by combining the features at each resolution in a recursive manner. In our experiments with deep saliency techniques, we use two state-of-the-art networks, PiCA-Net \cite{picanet} and $R^3$-Net \cite{r3net} (explained in Section \ref{subsubsec:deep-saliency}), to generate saliency maps from thermal images and to benchmark our salient pedestrian detection dataset. 

\section{Approach}
\label{sec:approach}
In this section, we explain the task of pedestrian detection in thermal images using Faster R-CNN \cite{ren2015faster}. We then present our proposed method of augmenting thermal images with their saliency maps to improve detection performance. Finally, we describe our motivation and efforts at annotating a subset of the KAIST Multispectral Pedestrian dataset \cite{kaist_benchmarkpaper} at the pixel level, for use by deep saliency networks.

\begin{figure*}[t]
\centering
\begin{tabular}{cccc}
    \includegraphics[height=275pt, width=0.33\linewidth]{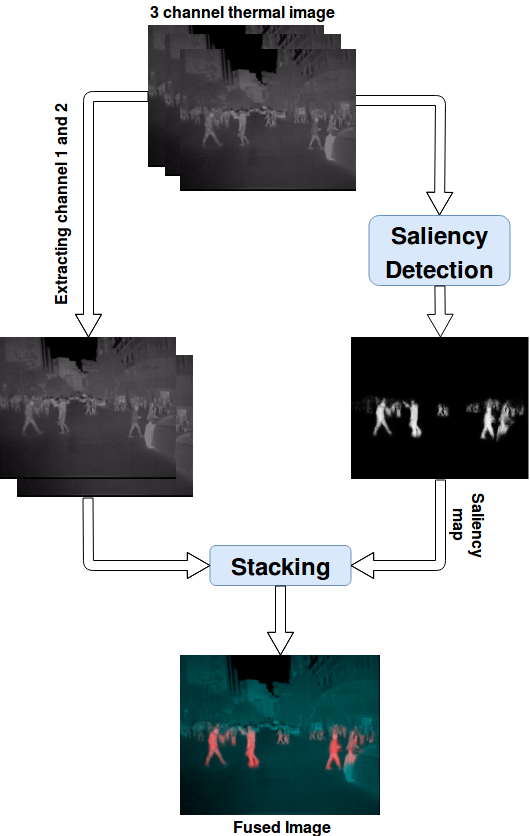} & & &
    \includegraphics[height=275pt, width=0.5\linewidth]{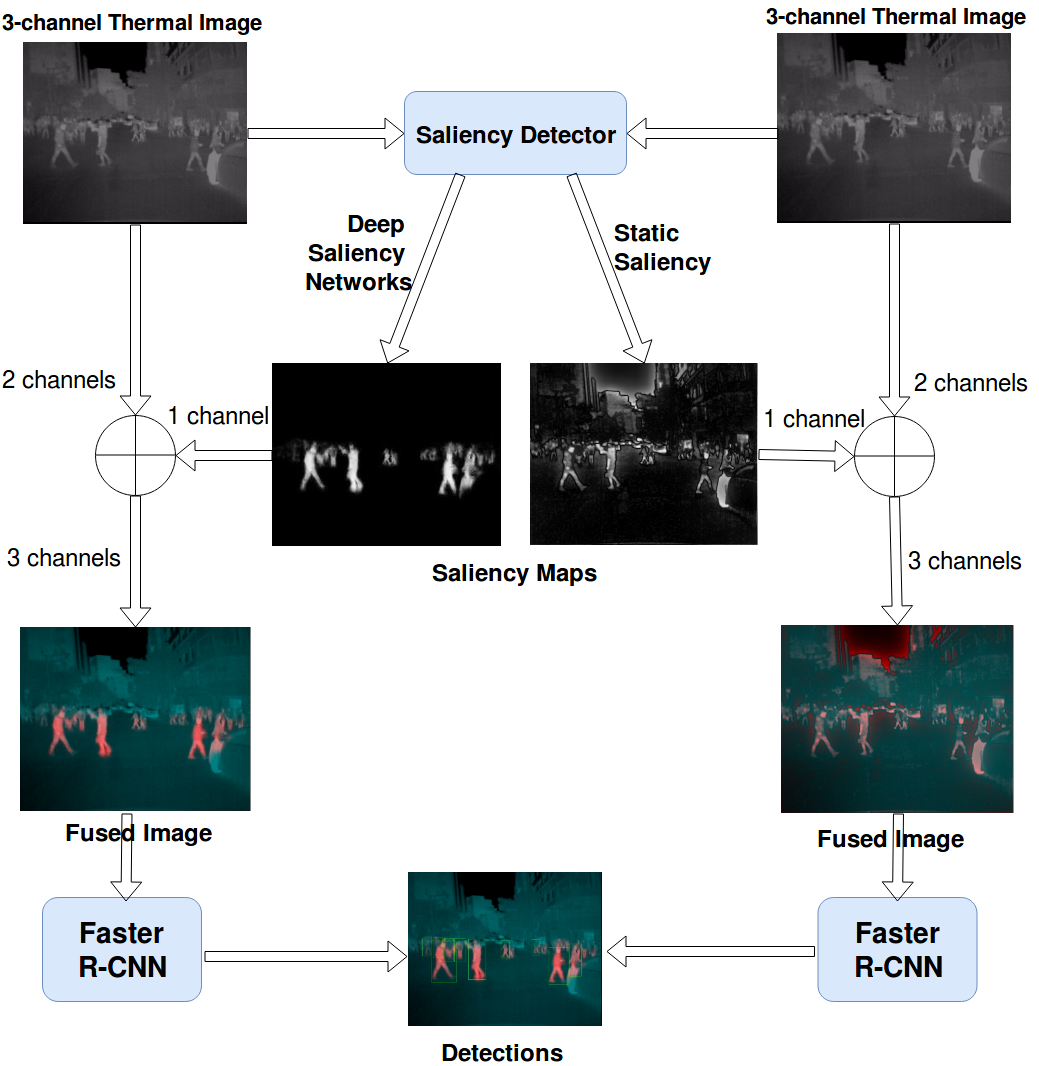}\\
    (a) & & & (b) 
\end{tabular}
\caption{\textbf{(a)} Procedure for augmenting thermal images with saliency maps, \textbf{(b)} Faster R-CNN training procedure on augmented images}
\label{fig:block-diagram}
\end{figure*}

\subsection{Baseline for Pedestrian Detection in Thermal Images using Faster R-CNN}
\label{subsec:baseline}
We adapt the Faster R-CNN \cite{ren2015faster} object detector for the task of pedestrian detection in thermal images. The Faster R-CNN architecture consists of a Region Proposal Network (RPN) that is used to propose regions in an image that are most likely to contain an object, and a Fast R-CNN \cite{girshick2015fast} network that classifies the objects present in that region along with refining their bounding box coordinates. Both these networks operate on shared convolutional feature maps generated by passing the input image through a backbone network (typically VGG16 \cite{VGG16} or ResNet101 \cite{he2016deep}). We train the Faster R-CNN end-to-end on the thermal images from the KAIST Multispectral Pedestrian dataset \cite{kaist_benchmarkpaper} and present the results in Table~\ref{tab:all-results}.

\subsection{Our Approach: Using Saliency Maps for Improving Pedestrian Detection}
\label{subsec:saliency}
\begin{figure*}[htb]
\centering
\setlength\tabcolsep{2pt} 
\begin{tabular}{ccccc}
     \includegraphics[width=0.18\linewidth]{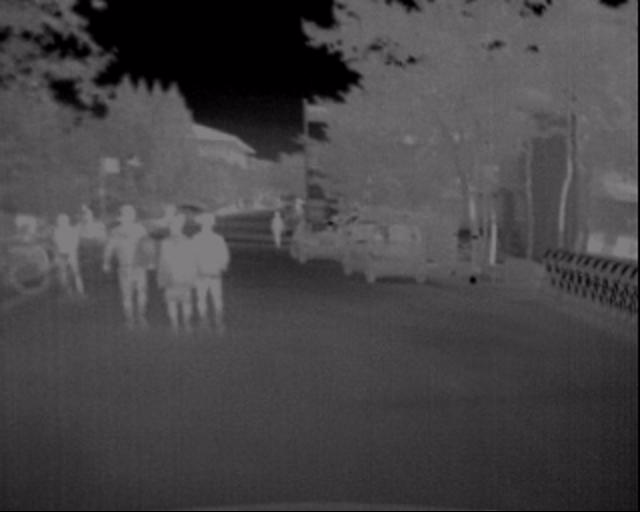} &
     \includegraphics[width=0.18\linewidth]{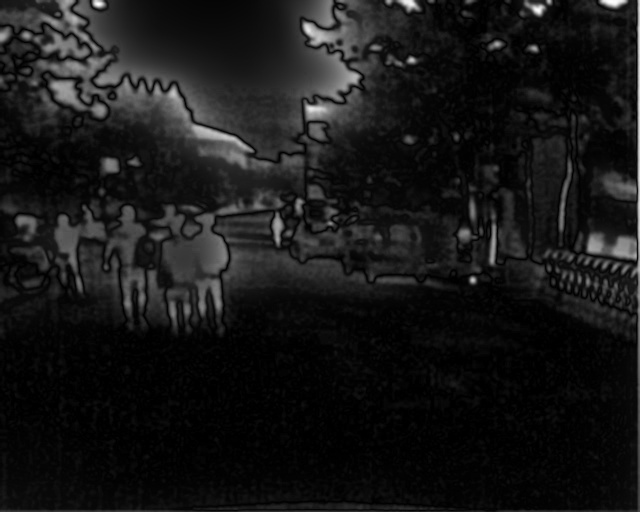} &
     \includegraphics[width=0.18\linewidth]{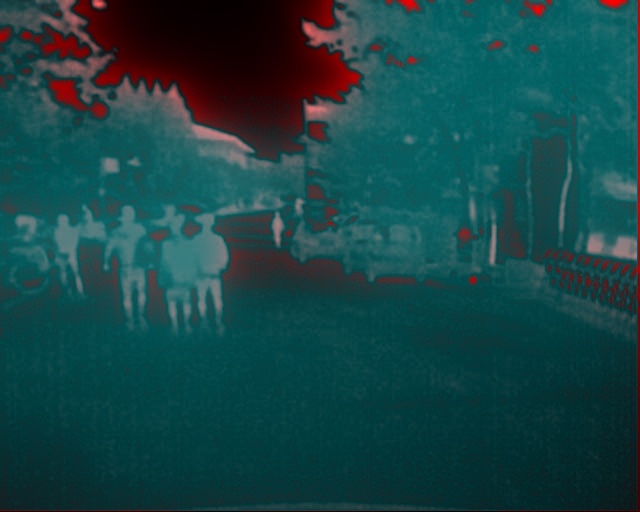} &
     \includegraphics[width=0.18\linewidth]{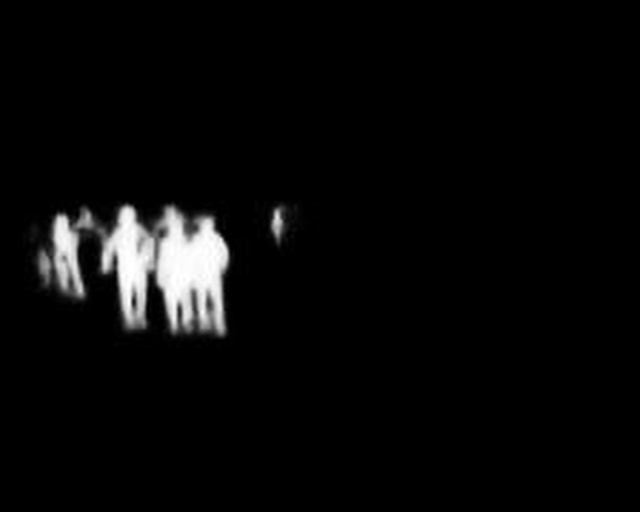} &
     \includegraphics[width=0.18\linewidth]{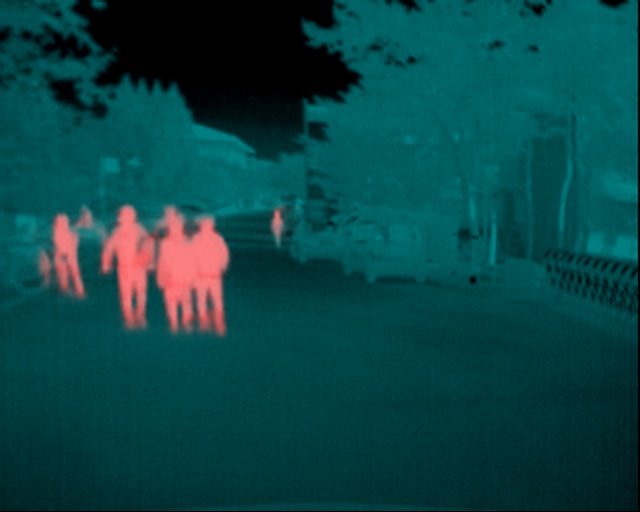} \\
     \includegraphics[width=0.18\linewidth]{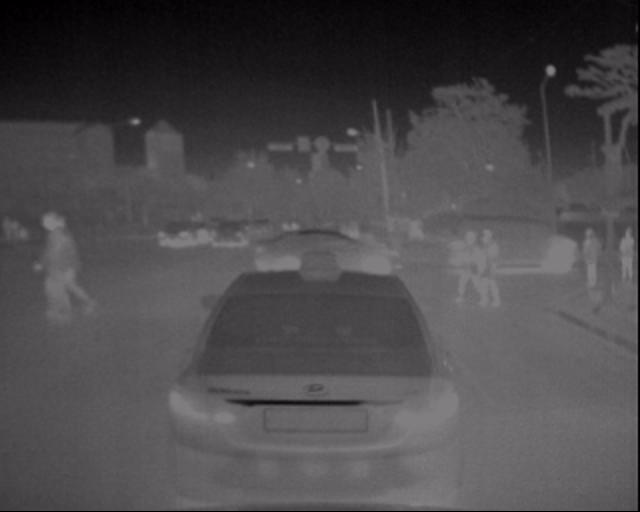} &
     \includegraphics[width=0.18\linewidth]{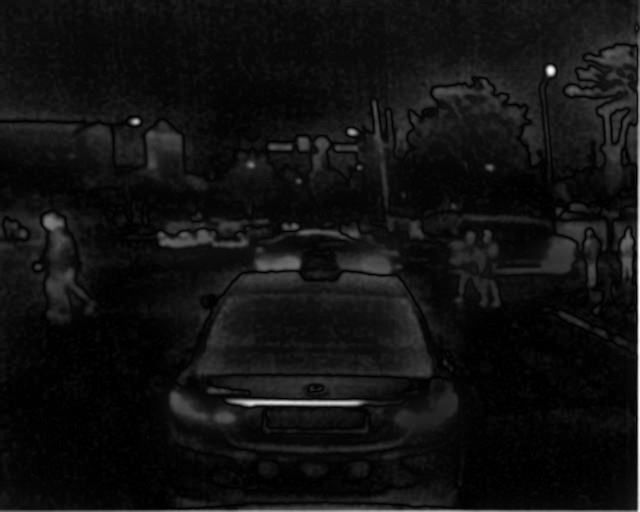} &
     \includegraphics[width=0.18\linewidth]{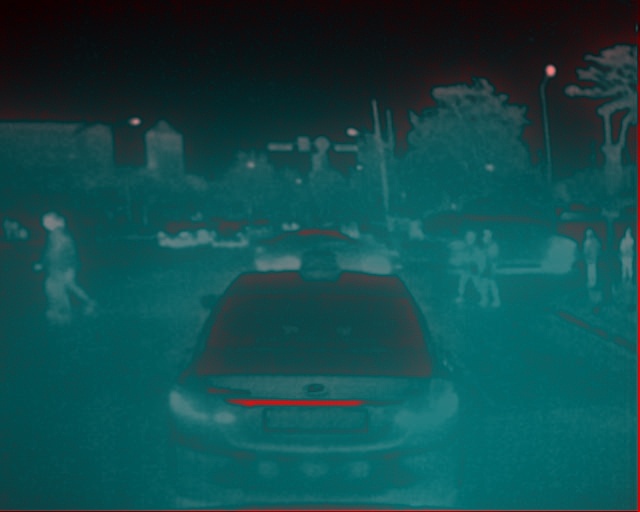} &
     \includegraphics[width=0.18\linewidth]{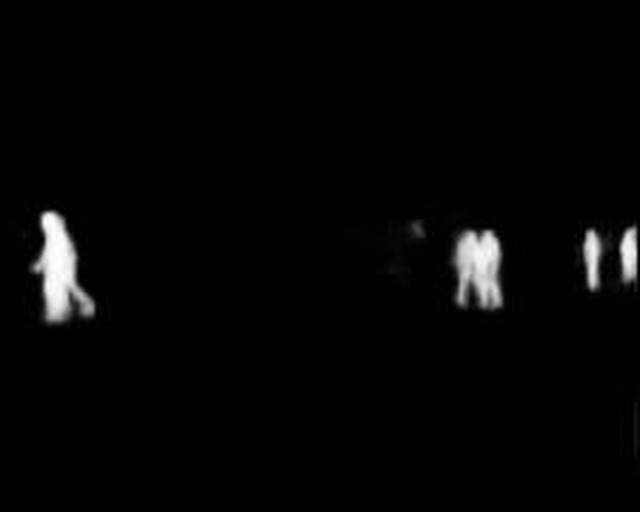} &
     \includegraphics[width=0.18\linewidth]{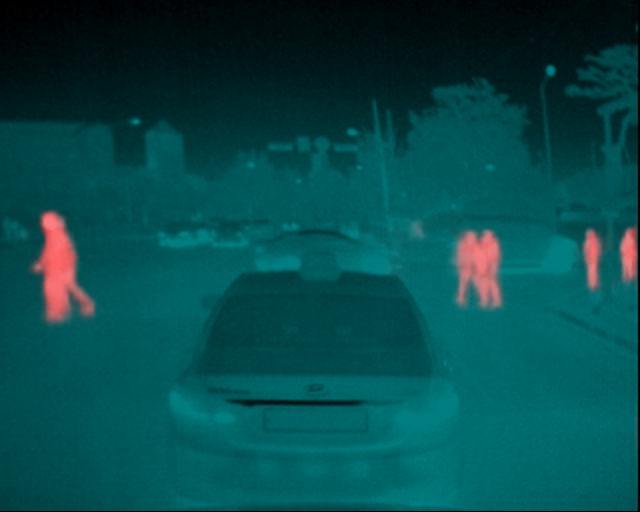} \\
     \includegraphics[width=0.18\linewidth]{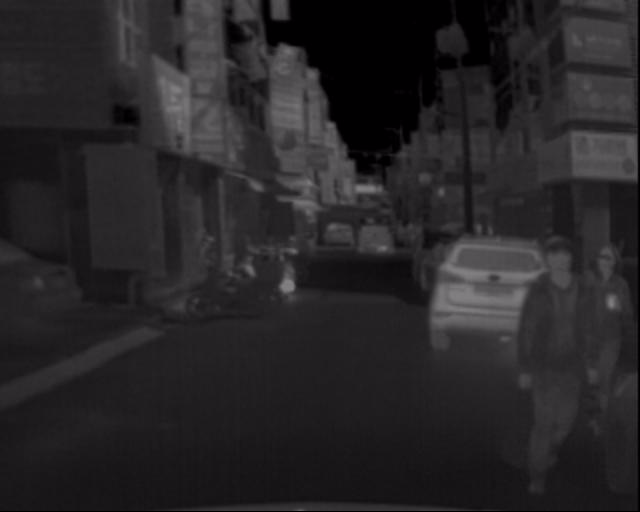} &
     \includegraphics[width=0.18\linewidth]{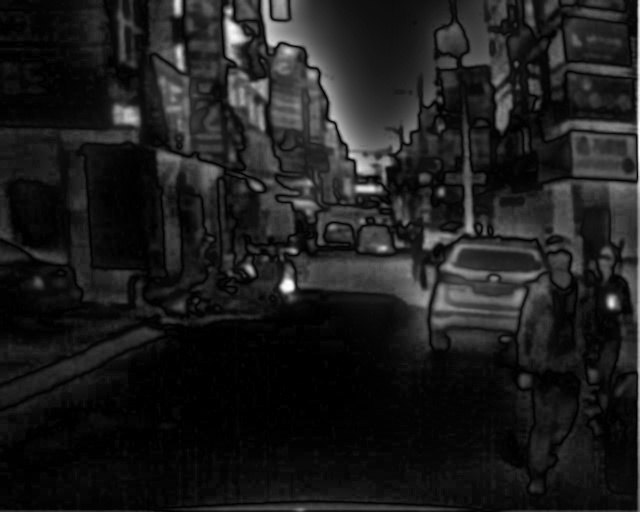} &
     \includegraphics[width=0.18\linewidth]{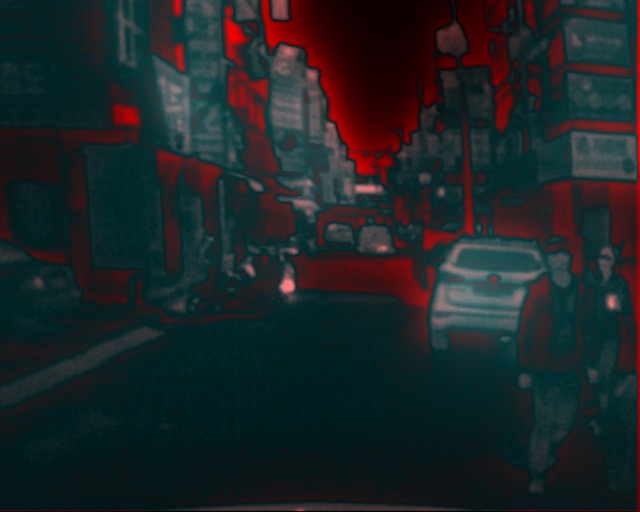} &
     \includegraphics[width=0.18\linewidth]{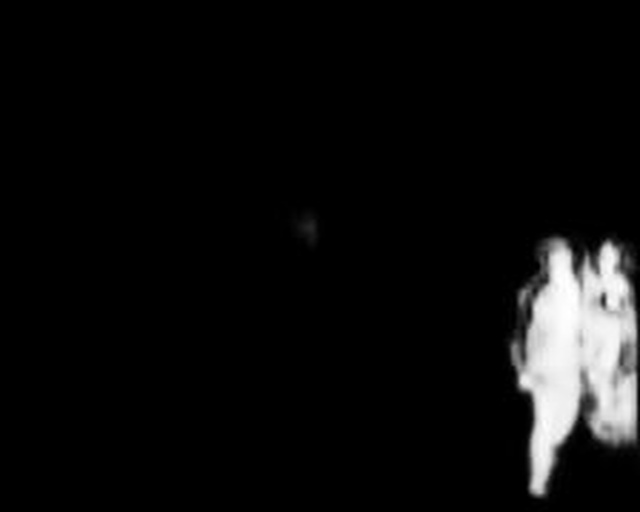} &
     \includegraphics[width=0.18\linewidth]{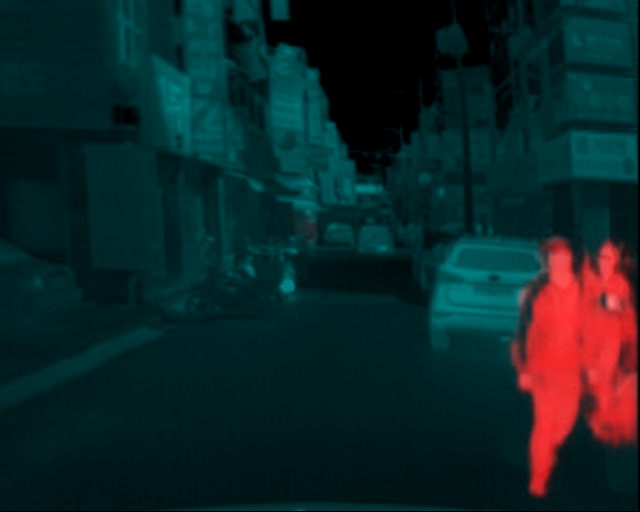} \\
     \includegraphics[width=0.18\linewidth]{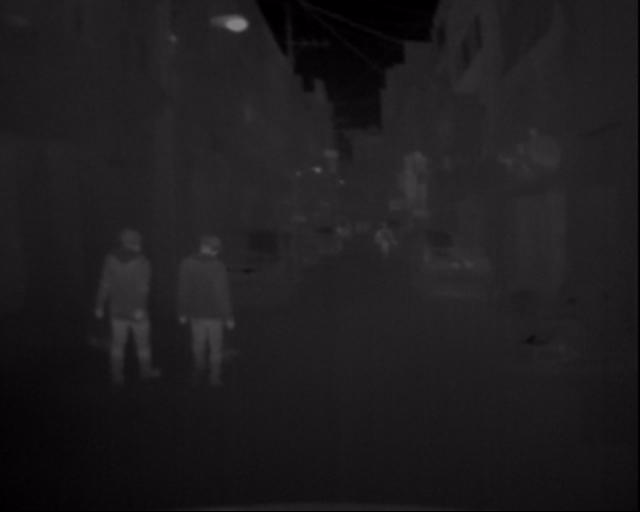} &
     \includegraphics[width=0.18\linewidth]{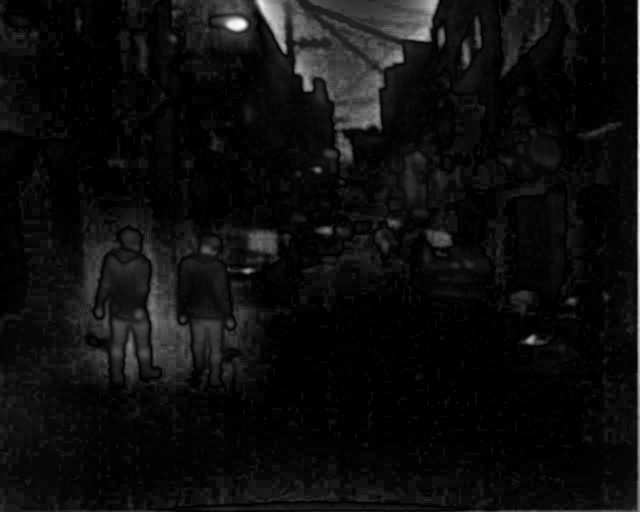} &
     \includegraphics[width=0.18\linewidth]{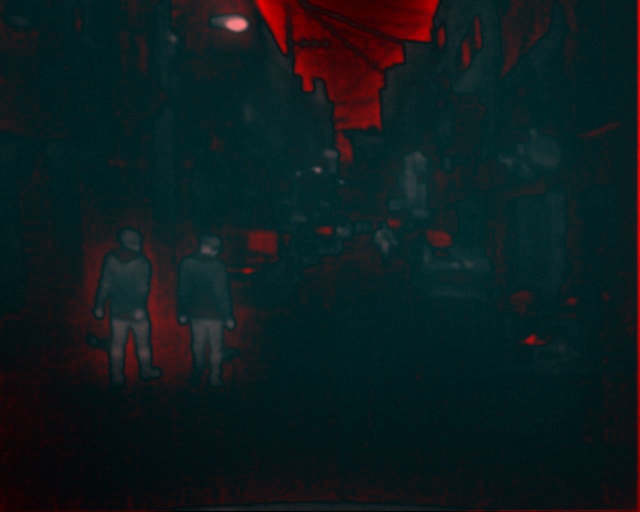} &
     \includegraphics[width=0.18\linewidth]{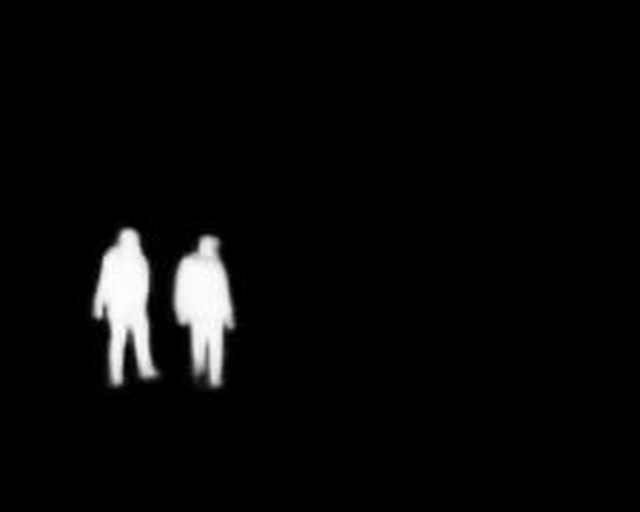} &
     \includegraphics[width=0.18\linewidth]{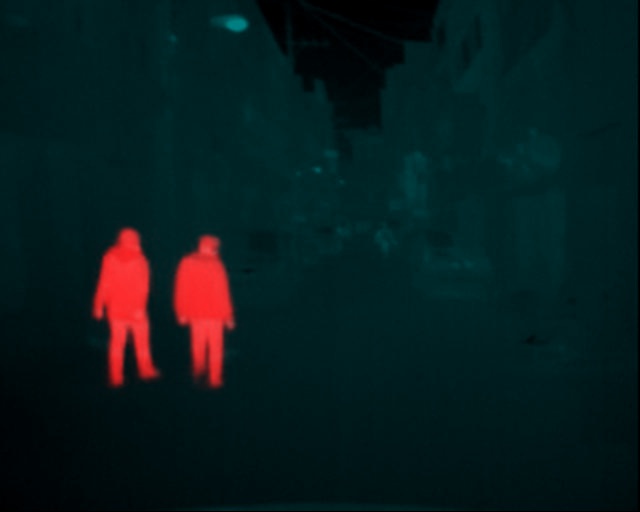} \\
     (a) & (b) & (c) & (d) & (e) \\
\end{tabular}
\caption{\textbf{Thermal images and generated saliency maps} for day (top 2 rows) and night (bottom 2 rows) images from the test set. (a) Original thermal images, (b) Static saliency maps, (c) Thermal images fused with static saliency maps, (d) Deep saliency maps, (e) Thermal images fused with deep saliency maps}
\label{fig:sal_viz}
\end{figure*}

We propose to use saliency maps extracted from thermal images in order to teach the pedestrian detector to ``see" better through pixel level context. We expect that such a system would perform better especially during daytime when humans are more indiscernible from their surroundings in thermal images. 
However, saliency maps discard all textural information available in thermal images. In order to mitigate this, we augment the thermal images with their saliency maps. We do this by replacing one duplicate channel of the 3-channel thermal images with the corresponding saliency maps as shown in Figure \ref{fig:block-diagram}(a). As seen in Figure \ref{fig:sal_viz}, the combination of saliency maps with thermal images help illuminate the salient parts of the image, while retaining the textural information in the image. As shown in Figure \ref{fig:block-diagram}(b), we then proceed to train the Faster R-CNN described in Section~\ref{subsec:baseline} on (i) saliency maps extracted from thermal images and (ii) thermal images augmented with their saliency maps generated using the two approaches described below.

\subsubsection{Static Saliency}

In this paper, we generate static saliency maps using OpenCV library \cite{opencv_library} that uses methods described in  \cite{Hou07saliencydetection:} and \cite{montabone2010human}. However, the saliency maps generated using this na{\"i}ve method highlight not only pedestrians but also other salient objects in the image (as seen in Figure \ref{fig:sal_viz} (b) \& (c)). This leaves room for a more powerful saliency detection technique that would highlight only the salient pedestrians and not any other salient objects in the image.

\subsubsection{Deep Saliency Networks}
\label{subsubsec:deep-saliency}
We investigate two state-of-the-art deep saliency networks in this paper.
\par\noindent
\textbf{PiCA-Net} \cite{picanet} is a pixel-wise contextual attention network which generates an attention map for each pixel corresponding to its relevance at each location. It uses a Bidirectional LSTM to scan the image horizontally and vertically  around a pixel to obtain its global context. For the local context, the attention operation is performed on a local neighboring region using convolutional layers. Finally a U-Net architecture is used to integrate the PiCA-Nets hierarchically for salient object detection. 
\par\noindent
\textbf{$\mathbf{R^3}$ Net} \cite{r3net} uses a Residual Refinement Block (RRB) to learn the residuals between the ground truth and the saliency map in a recursive manner. The RRB alternatively utilizes low-level features and high-level features to refine the saliency maps at each recurrent step by adding the previous saliency map to the learned residual. 
\par\noindent
As seen in Figure~\ref{fig:sal_viz} (d) \& (e), these techniques illuminate only the pedestrians in a scene.

\subsection{Our Dataset: Annotating KAIST Multispectral Pedestrian for Salient Pedestrian Detection}
\label{subsec:salient-dataset}
\begin{figure*}[h]
\centering
\setlength\tabcolsep{2pt} 
\begin{tabular}{cccc}
\centering
\includegraphics[width=0.2\linewidth]{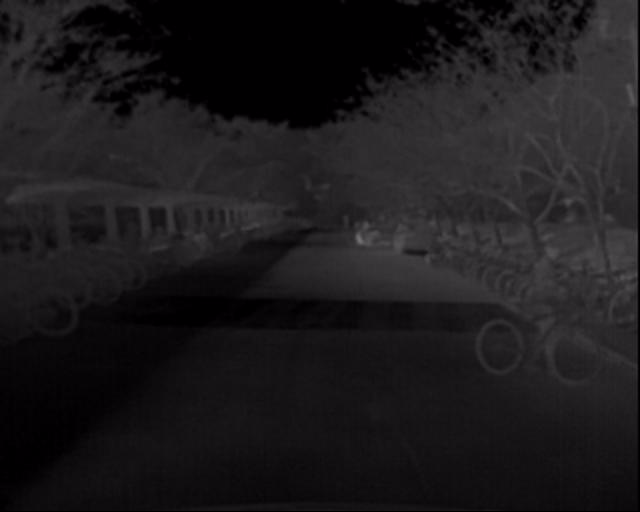} & 
\includegraphics[width=0.2\linewidth]{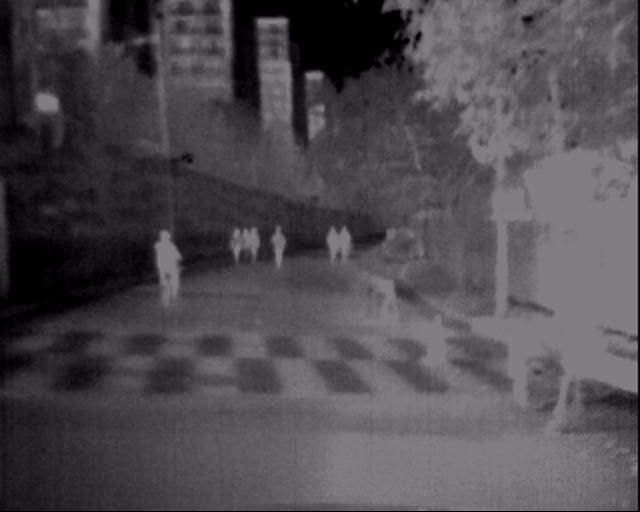} &
\includegraphics[width=0.2\linewidth]{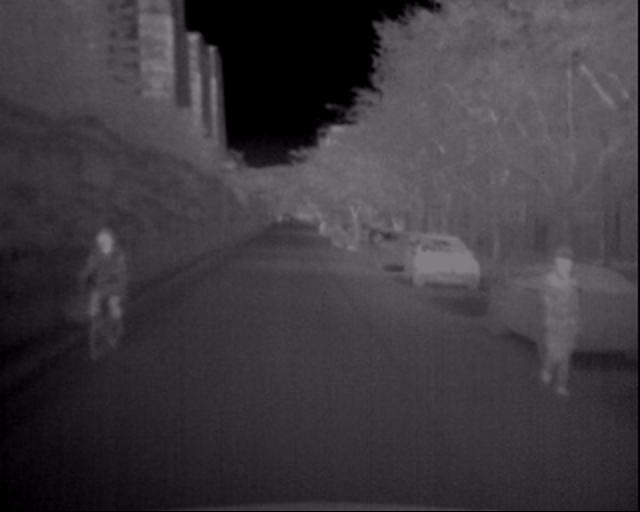} & 
\includegraphics[width=0.2\linewidth]{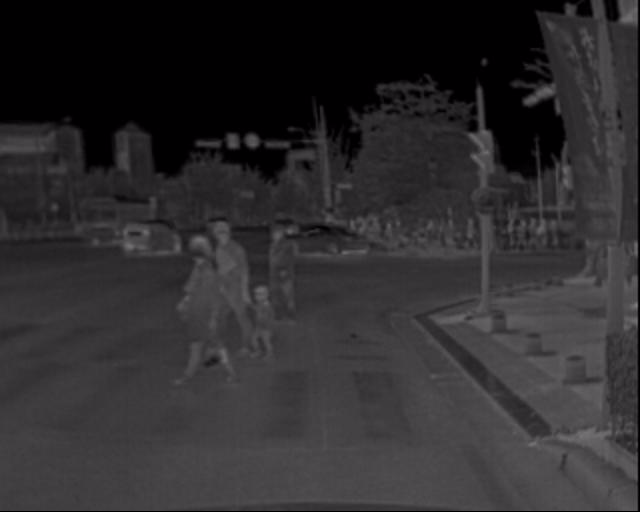} \\ 
\includegraphics[width=0.2\linewidth]{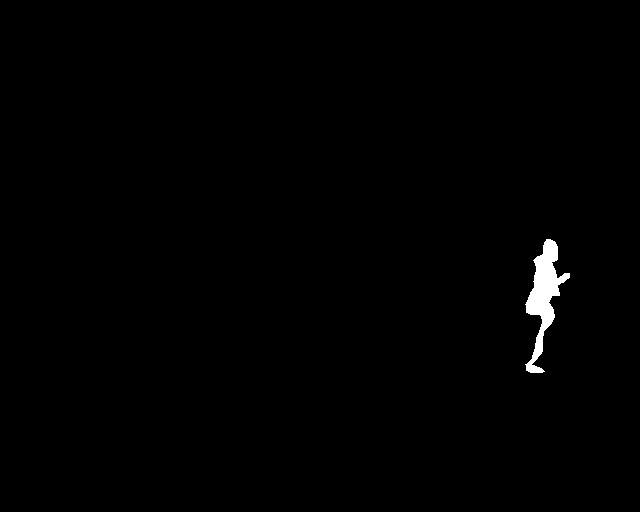} &  
\includegraphics[width=0.2\linewidth]{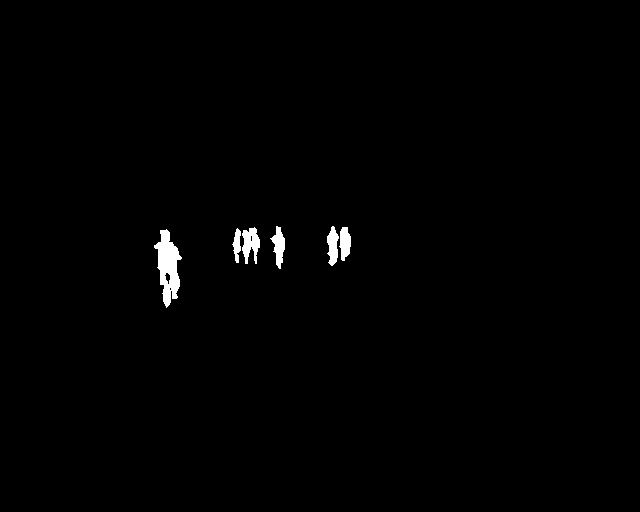} &
\includegraphics[width=0.2\linewidth]{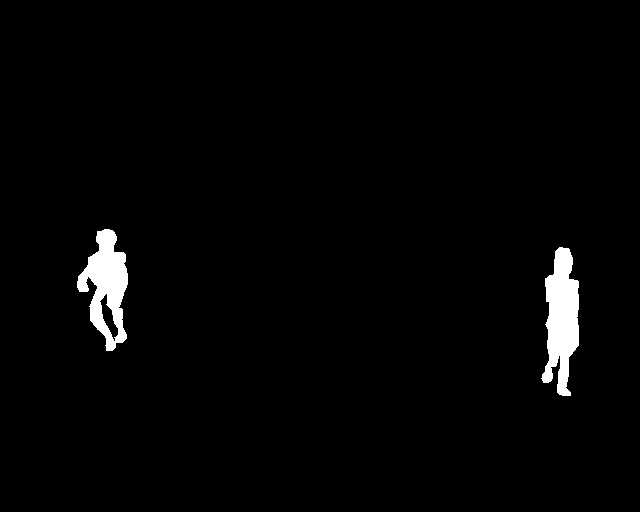} & 
\includegraphics[width=0.2\linewidth]{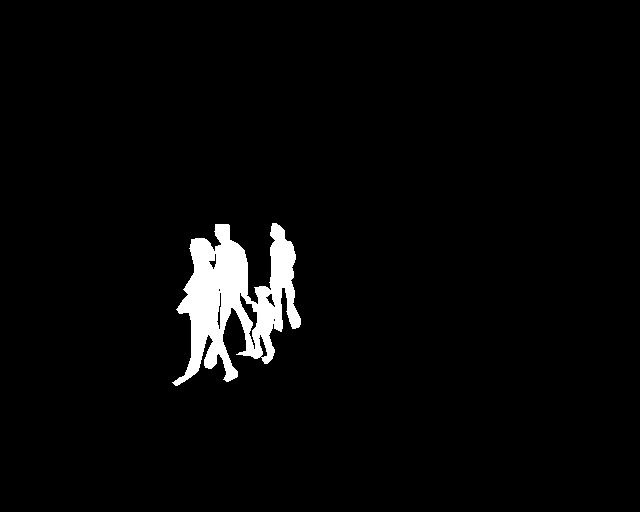}
\end{tabular}
\caption{\textbf{Sample annotations from our KAIST Pedestrian Saliency Dataset}. Top: Original images, Bottom: Pixel level annotations}
\label{fig:saliency-dataset}
\end{figure*}


\begin{figure*}[h]
\centering
\begin{tabular}{cc}
\includegraphics[width=0.40\linewidth]{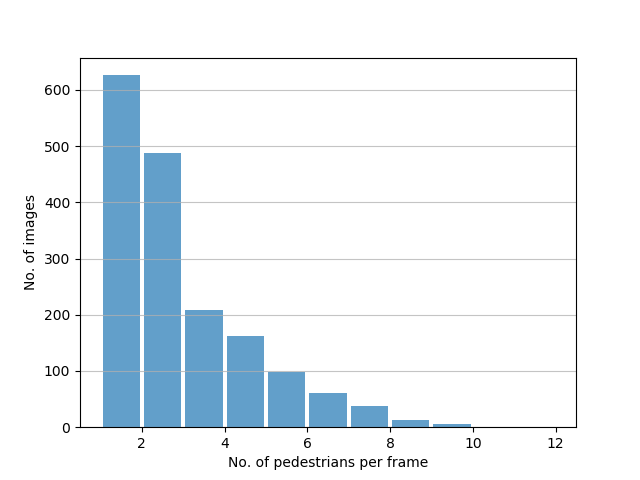} &  
\includegraphics[width=0.40\linewidth]{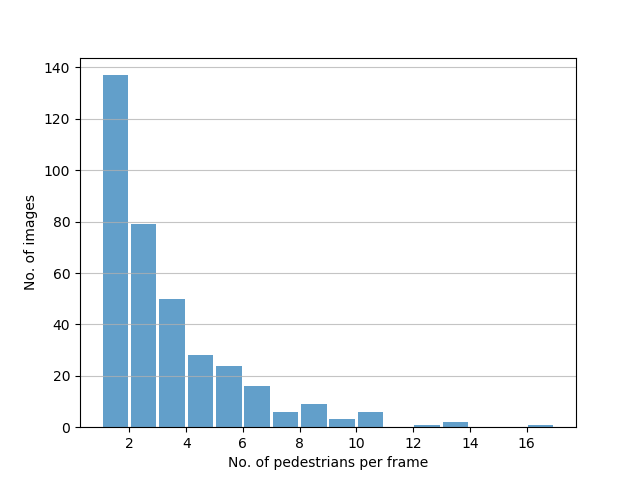} \\
(a) & (b)
\end{tabular}
\caption{\textbf{Distribution of Pedestrians} in (a) training images (b) test images}
\label{fig:data_stats}
\end{figure*}

In order to train a deep saliency network, we need pixel level annotations for salient objects. Since there are no publicly available thermal datasets with ground truth saliency masks for pedestrians, we create a pedestrian saliency dataset and make it publicly available \footnote{https://information-fusion-lab-umass.github.io/Salient-Pedestrian-Detection/} to facilitate further research on the use of saliency techniques for multispectral pedestrian detection.

We select 1702 images from the training set of the KAIST Multispectral Pedestrian dataset \cite{kaist_benchmarkpaper}, by sampling every $15^{th}$ image from all images captured during day and every $10^{th}$ image from all images captured during night, containing pedestrians. These images were selected in order to have roughly the same number of images captured at both times of the day (913 day images and 789 night images), containing 4170 instances of pedestrians. We manually annotate these images using the VGG Image Annotator \cite{dutta2016via} tool to generate the ground truth saliency masks based on the location of the bounding boxes on pedestrians in the original dataset. Additionally, we create a set of 362 images with similar annotations from the test set to validate our deep saliency detection networks, with 193 day images and 169 night images, containing 1029 instances of pedestrians. Figure \ref{fig:saliency-dataset} shows sample images and annotations from the new KAIST Pedestrian Saliency Detection Dataset. The distribution of pedestrians per frame in the training and test sets are shown in Figure \ref{fig:data_stats}. Note however that the pixel level annotations are not completely precise, so these annotations might not be suitable for fine semantic segmentation tasks. However, benchmark results in Table~\ref{tab:deep-saliency-results} show that this dataset works reasonably well for salient pedestrian detection tasks. 

\section{Experiments}
\label{sec:exp}


\subsection{Datasets and Evaluation Protocols}
\label{sec:dataset}
For training the pedestrian detectors, we use the thermal images from the KAIST Multispectral Pedestrian Dataset \cite{kaist_benchmarkpaper} that contains approximately $50k$ training images and $45k$ test images from videos captured during different times of the day using thermal and RGB cameras. Following the evaluation protocol in \cite{liu2017exploiting, liu2016multispectral}, we sample images every 3 frames from training videos and every 20 frames from test videos, and exclude occluded, truncated, and small ($<50$ pixels) pedestrian instances. This gives us 7,601 training images (4,755 day, 2,846 night) and 2,252 test images (1,455 day, 797 night). We use the improved annotations for these 2,252 test images given in \cite{liu2016multispectral}. For training deep saliency networks, we annotate a subset of the KAIST Multispectral Pedestrian dataset as described in Section~\ref{subsec:salient-dataset}. Once the deep saliency networks are trained, we use them to generate saliency maps for the 7,601 training and 2,252 test images and these are then used to augment the thermal images as described in Section~\ref{subsec:saliency}.

For evaluating pedestrian detection, we report the Log Average Miss Rate (LAMR) over the range [$10^{-2}, 10^0$] against the False Positives Per Image (FPPI) under reasonable conditions \cite{dollar2012pedestrian} for day and night images. We also report the mean Average Precision (mAP) of detections at IOU=$0.5$ with the ground truth box. For evaluation of saliency detection, we use two metrics - F-measure score ($F_\beta$) which is a weighted harmonic mean of the precision and recall, and Mean Absolute Error (MAE) which computes the average absolute per pixel difference between predicted saliency maps and corresponding ground truth saliency maps \cite{hou2017deeply}.

\subsection{Implementation Details}
%
\subsubsection{Faster R-CNN for Pedestrian Detection}
We use an open source implementation \cite{jwyang-code} of the original Faster R-CNN network with a few modifications.
First, we remove the fifth max-pooling layer of the VGG16 backbone network. The original Faster R-CNN used 3 scales and 3 ratios for the reference anchors. We use 9 scales for the reference anchors, between 0.05 and 4. The Faster R-CNN network is initialized with VGG16 weights pretrained on ImageNet\cite{imagenet} and fine-tuned on data sources described in Section~\ref{subsec:saliency} for 6 epochs. We fix the first two convolutional layers of the VGG16 model and fine-tune the rest using SGD with momentum of 0.9, learning rate of 0.001, batch size of 1, and train our model using two NVIDIA Titan X GPUs with 12GB memory each. 

\subsubsection{Deep Saliency Networks}
We train PiCA-Net \cite{picanet} and $R^3$-Net \cite{r3net} on thermal images with pixel level annotations.
For PiCA-Net, we use an open source implementation \cite{yoo2018picanet} and keep the same network architecture as described in the original paper. For training, we augment the training images with random mirror-flipping and random crops. The decoder is trained from scratch with a learning rate of $0.01$ and encoder is fine-tuned with a learning rate of $0.001$ for $16$ epochs and decayed by $0.1$ for another $16$ epochs. We used SGD optimizer with momentum $0.9$ and weight decay $0.0005$. The entire setup is trained with a batch size of $4$ on a single NVIDIA GTX 1080ti GPU. Also, since the generated saliency maps are of size $224\times224$, we resize it to the original image size using Lanczos interpolation \cite{turkowski1990filters}.
For $R^3$-net we use the authors' implementation. As described, we initialize the parameters of the feature extraction network using weights from the ResNeXt \cite{resnext} network. We use SGD with learning rate $0.001$, momentum $0.9$, weight decay $0.0005$ and train for $9000$ iterations using batch size of $10$ on two NVIDIA Titan X GPUs with 12GB memory each. 

\subsection{Results and Analysis}

\subsubsection{Performance of Deep Saliency Networks on our KAIST Salient Pedestrian Detection dataset}
\label{subsec:benchmark}
We evaluate the performance of the PiCA-Net and $R^3$-Net on the test set of our annotated KAIST Salient Pedestrian Detection dataset to provide a benchmark. The results are summarized in Table~\ref{tab:deep-saliency-results} and show reasonable saliency detection performance. Saliency masks generated using these networks can be seen in Figure~\ref{fig:sal_viz} (d) \& (e). Note that the saliency maps generated from the $R^3$-Net have been post-processed using a fully-connected CRF \cite{krahenbuhl2011efficient} to improve coherence, resulting in the slightly better results as compared to PiCA-Net.

\begin{table}[h]
    \centering
    \begin{tabular}{|c|c|c|}
        \hline
         \textbf{Method} & \textbf{$\mathbf{F_\beta}$ score} & \textbf{MAE}  \\
         \hline\hline
         PiCA-Net & 0.5942 & 0.0062\\
         \hline
         $R^3$-Net & 0.6417 & 0.0049\\
         \hline
    \end{tabular} 
    \caption{\textbf{Performance of deep saliency networks} on our annotated test set}
    \label{tab:deep-saliency-results}
\end{table}

\subsubsection{Quantitative analysis of Pedestrian Detection in Thermal Images using Saliency Maps}

\begin{figure}[h]
    \centering
    \includegraphics[width=\linewidth]{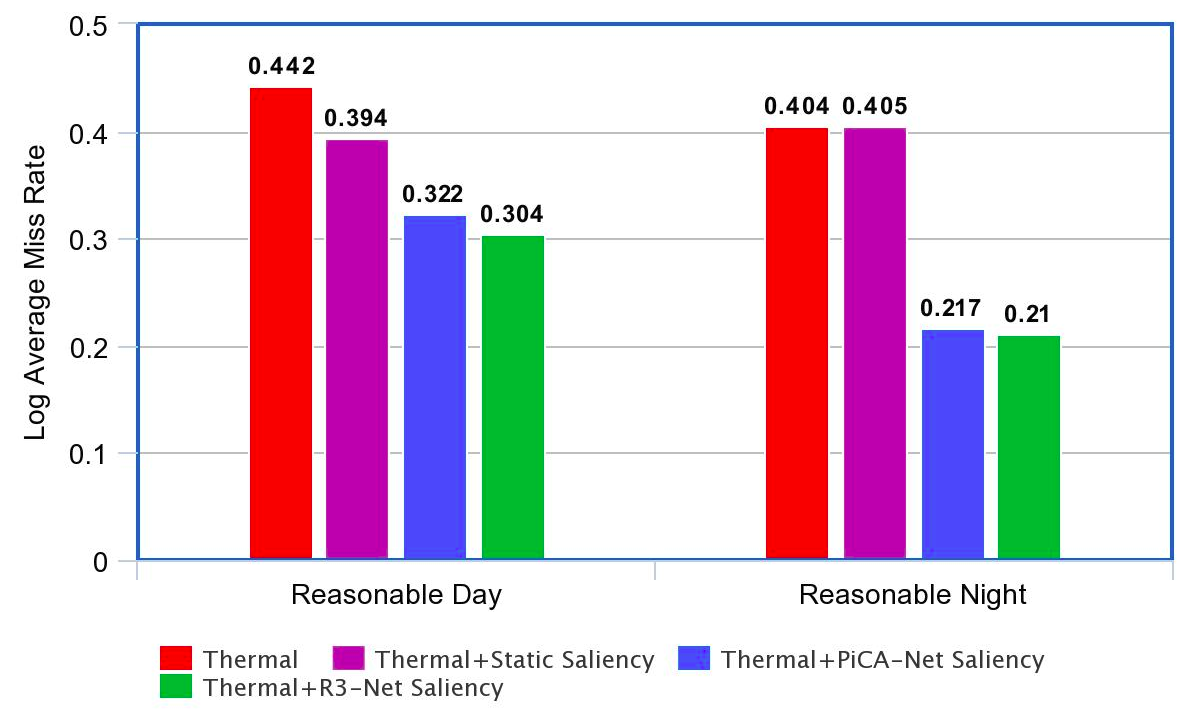}
    \caption{\textbf{Comparison of Miss Rates from different models}}
    \label{fig:comparison}
\end{figure}

\begin{table*}[ht]
\centering
\begin{tabular}{|c|c|c|c|c|c|c|c|c|}
\hline
\multirow{2}{*}{\textbf{\begin{tabular}[c]{@{}c@{}}Testing\\ Condition\end{tabular}}} & \multirow{2}{*}{\textbf{Metric}} & \multicolumn{7}{c|}{\textbf{Dataset Used}}                                                                                                                                                                                                                                                                                                                                                                                                                                                                          \\ \cline{3-9} 
                                                                                      &                                  & \textbf{Thermal} & \textbf{\begin{tabular}[c]{@{}c@{}}Static \\ Saliency \\ Maps\end{tabular}} & \textbf{\begin{tabular}[c]{@{}c@{}}Static \\ Saliency \\ + Thermal\end{tabular}} & \textbf{\begin{tabular}[c]{@{}c@{}}PiCA-Net\\ Saliency \\ Maps\end{tabular}} & \textbf{\begin{tabular}[c]{@{}c@{}}PiCA-Net \\ Saliency\\ + Thermal\end{tabular}} & \textbf{\begin{tabular}[c]{@{}c@{}}R$^3$-Net \\ Saliency\\ Maps\end{tabular}} & \textbf{\begin{tabular}[c]{@{}c@{}}R$^3$-Net \\ Saliency\\ + Thermal\end{tabular}} \\ \hline\hline
\multirow{2}{*}{\textbf{Day}}                                                         & \textbf{mAP}                     & 0.616            & 0.590                                                                       & \textbf{0.645}                                                                   & 0.571                                                                        & 0.640                                                                             & 0.576                                                                      & \textbf{0.685}                                                                  \\ \cline{2-9} 
                                                                                      & \textbf{LAMR}                    & 0.442            & 0.479                                                                       & 0.394                                                                            & 0.342                                                                        & \textbf{0.322}                                                                    & 0.352                                                                      & \textbf{0.304}                                                                  \\ \hline\hline
\multirow{2}{*}{\textbf{Night}}                                                       & \textbf{mAP}                     & 0.655            & 0.605                                                                       & 0.641                                                                            & 0.639                                                                        & \textbf{0.676}                                                                    & 0.585                                                                      & \textbf{0.732}                                                                  \\ \cline{2-9} 
                                                                                      & \textbf{LAMR}                    & 0.404            & 0.462                                                                       & 0.405                                                                            & 0.285                                                                        & \textbf{0.217}                                                                    & 0.320                                                                      & \textbf{0.210}                                                                  \\ \hline
\end{tabular}
\caption{\textbf{Comparison of results from different techniques.} Our deep saliency map fused thermal images surpass all approaches in mean Average Precision (mAP) and Log Average Miss Rate (LAMR). Top 2 results are in bold.}
\label{tab:all-results}
\end{table*}

\begin{figure*}[h]
\centering
\begin{tabular}{cc}
\includegraphics[width=0.45\linewidth]{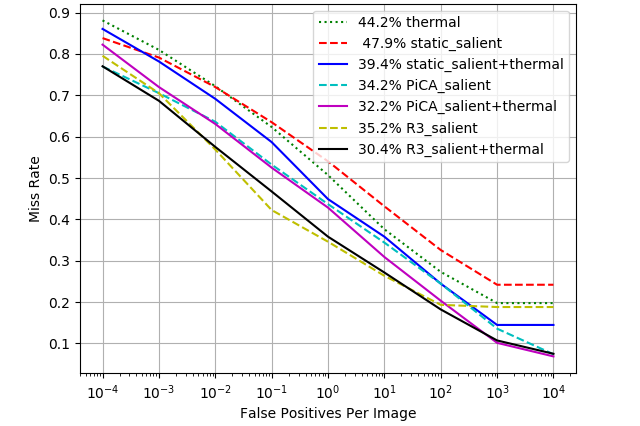} &  
\includegraphics[width=0.45\linewidth]{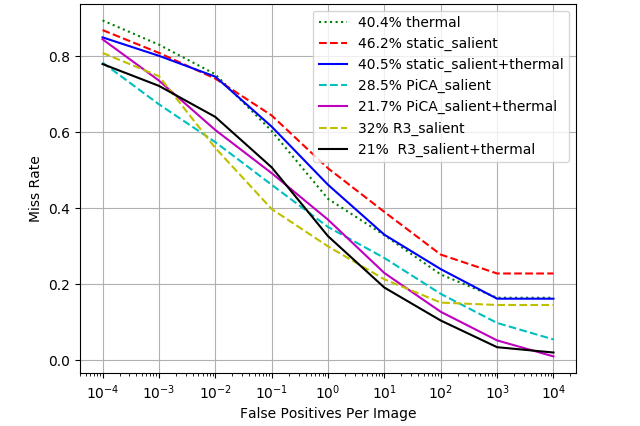} \\
(a) & (b)
\end{tabular}
\caption{\textbf{Miss Rate vs FPPI curves} for a) Day reasonable conditions b) Night reasonable conditions. Our deep saliency + thermal methods are the lower curves indicating better performance compared to baseline approaches.}
\label{fig:fppi_mr}
\end{figure*}

\begin{figure*}[htb]
\centering
\setlength\tabcolsep{2pt} 
\begin{tabular}{ccccc}
    & \multicolumn{2}{c}{\textbf{Day}} & \multicolumn{2}{c}{\textbf{Night}} \\
    (a) &
     \includegraphics[width=0.18\linewidth]{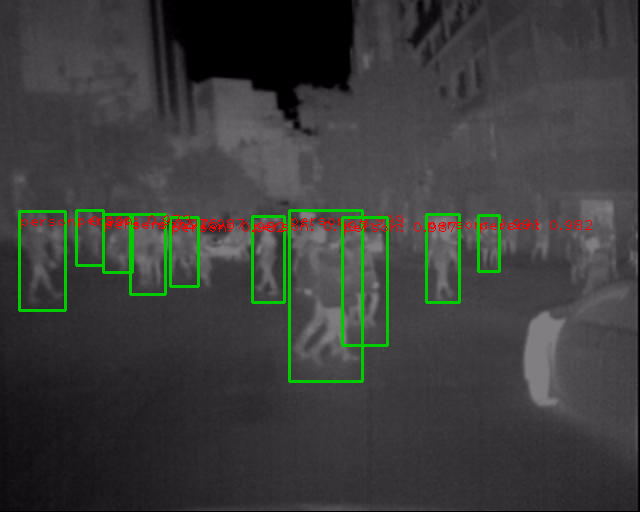} &
     \includegraphics[width=0.18\linewidth]{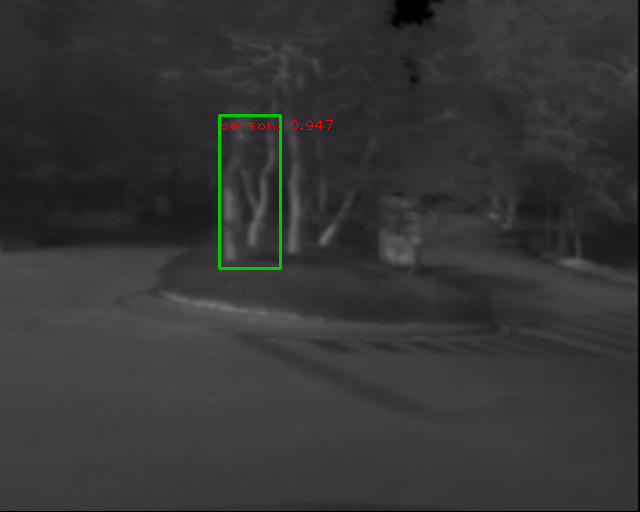} &
     \includegraphics[width=0.18\linewidth]{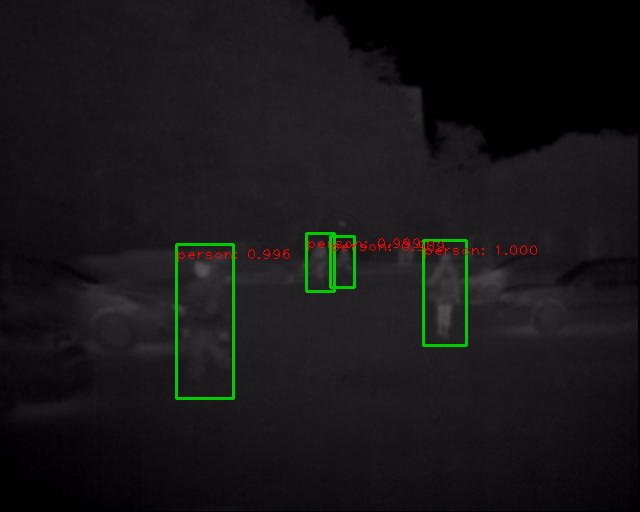} &
     \includegraphics[width=0.18\linewidth]{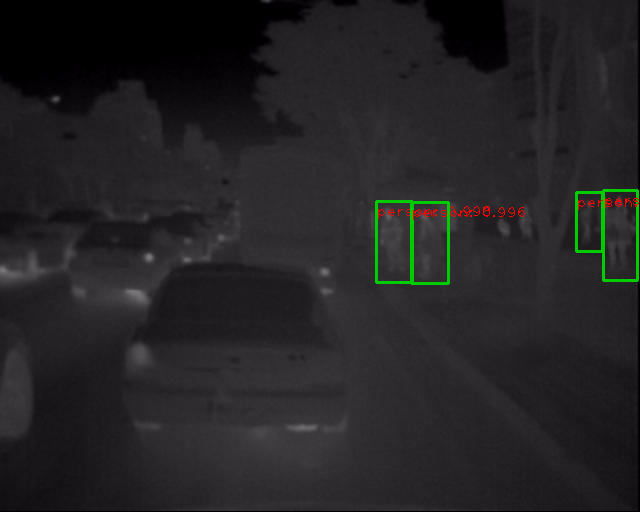} \\
     (b) &
     \includegraphics[width=0.18\linewidth]{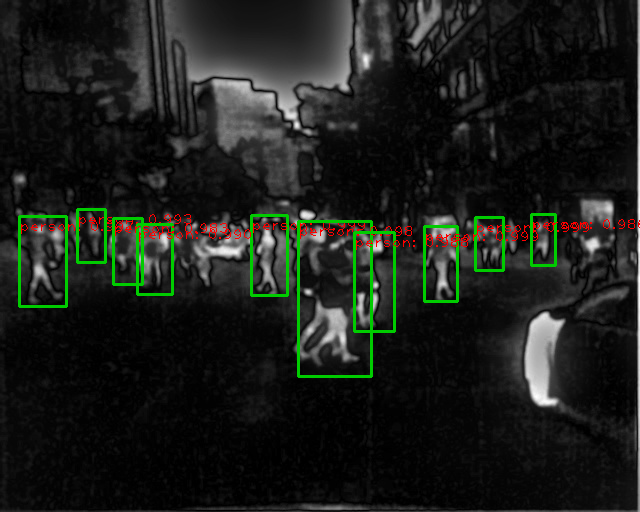} &
     \includegraphics[width=0.18\linewidth]{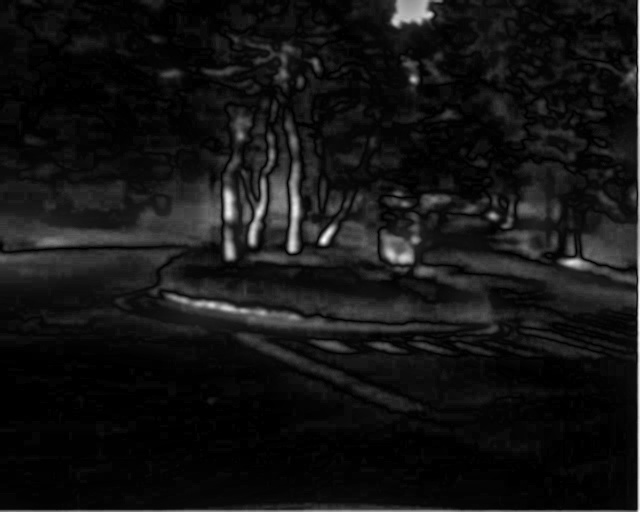} &
     \includegraphics[width=0.18\linewidth]{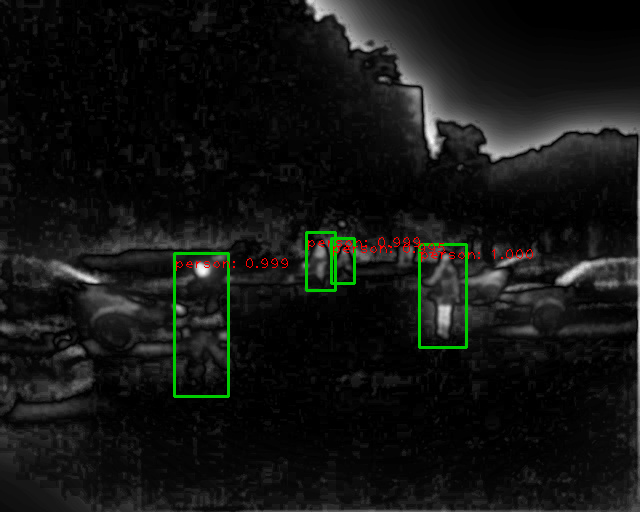} &
     \includegraphics[width=0.18\linewidth]{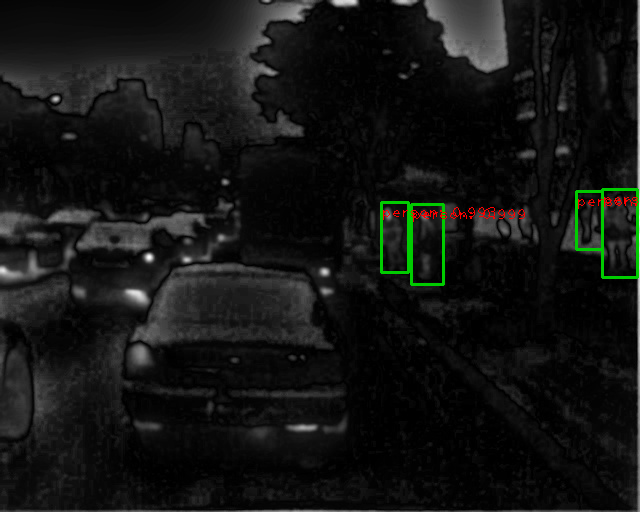} \\
     (c) &
     \includegraphics[width=0.18\linewidth]{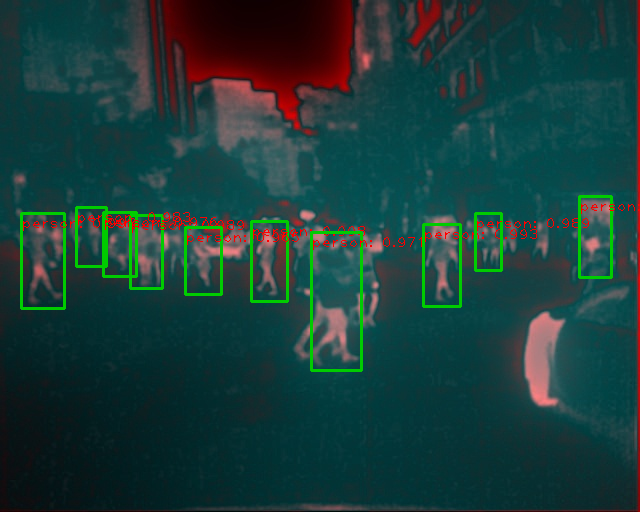} &
     \includegraphics[width=0.18\linewidth]{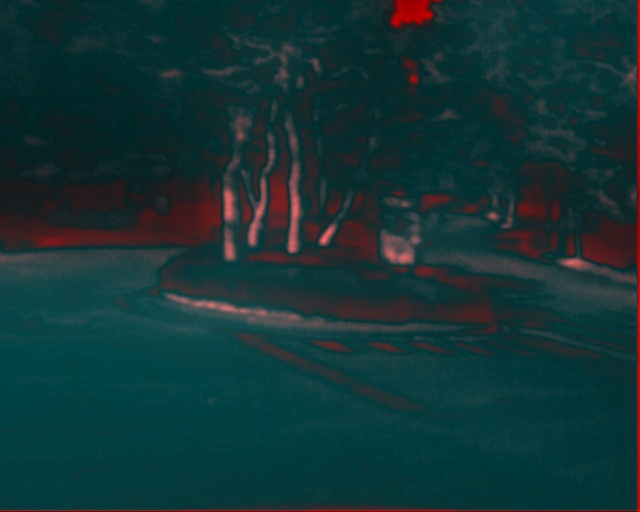} &
     \includegraphics[width=0.18\linewidth]{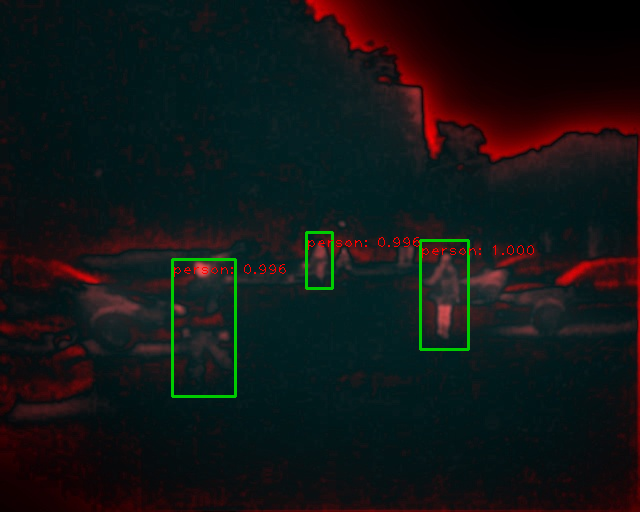} &
     \includegraphics[width=0.18\linewidth]{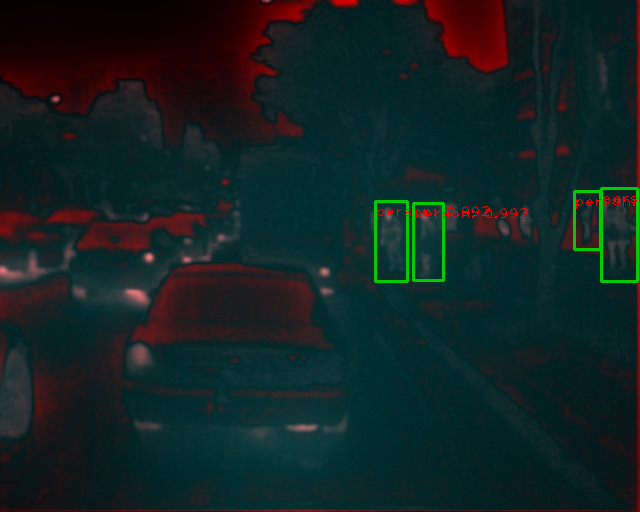} \\
     (d) &
     \includegraphics[width=0.18\linewidth]{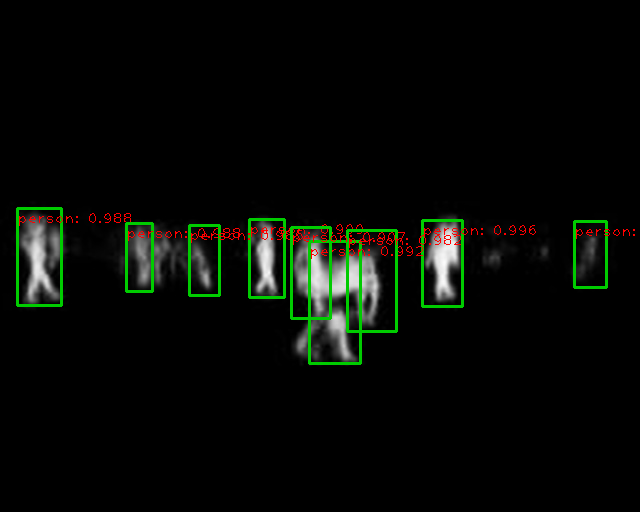} &
     \includegraphics[width=0.18\linewidth]{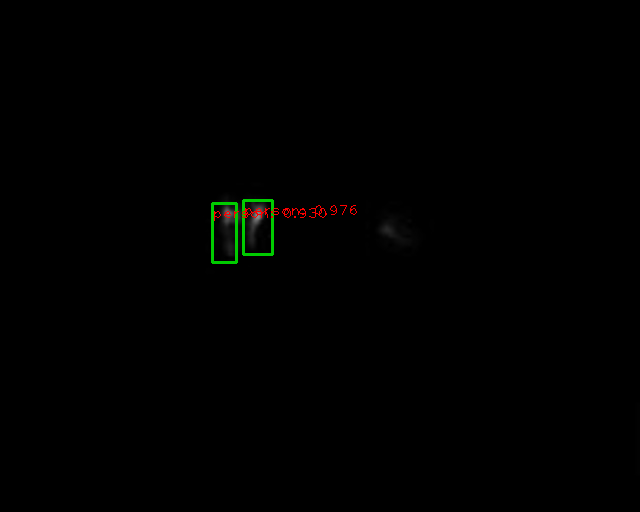} &
     \includegraphics[width=0.18\linewidth]{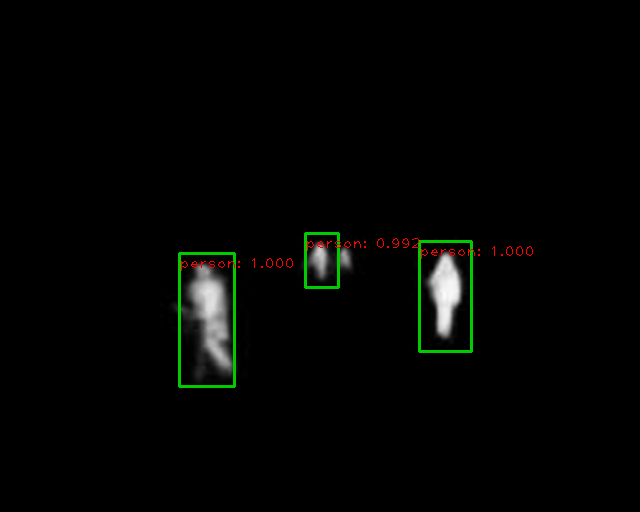} &
     \includegraphics[width=0.18\linewidth]{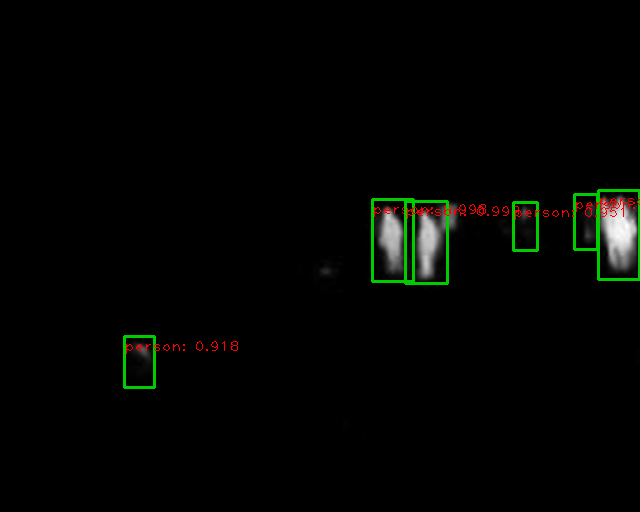} \\
     (e) &
     \includegraphics[width=0.18\linewidth]{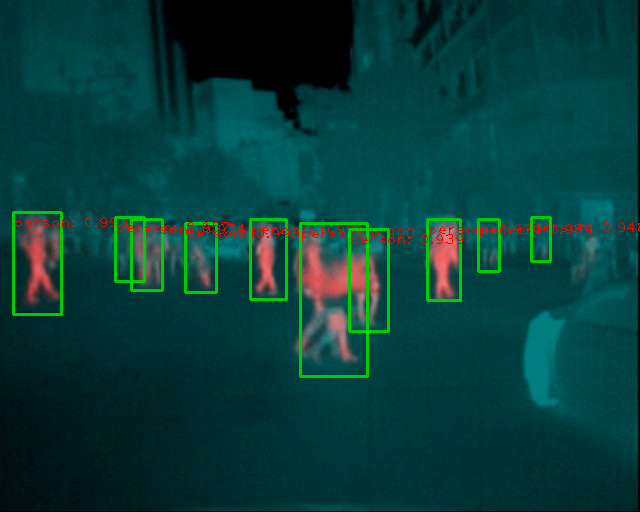} &
     \includegraphics[width=0.18\linewidth]{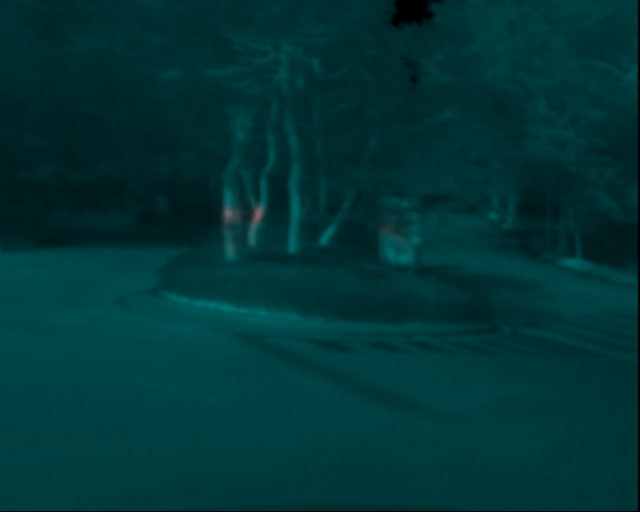} &
     \includegraphics[width=0.18\linewidth]{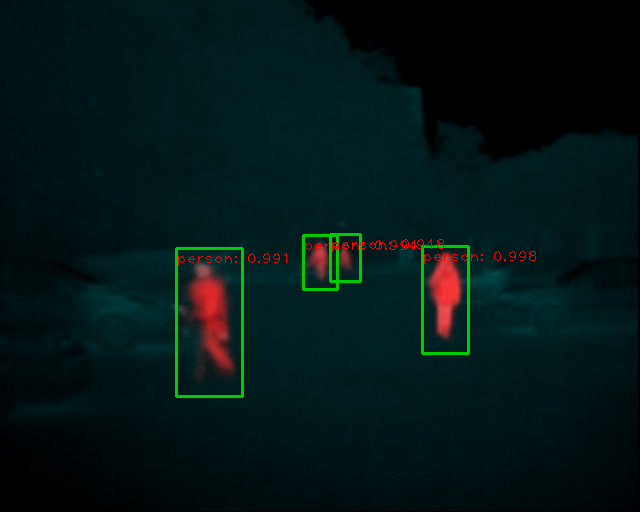} &
     \includegraphics[width=0.18\linewidth]{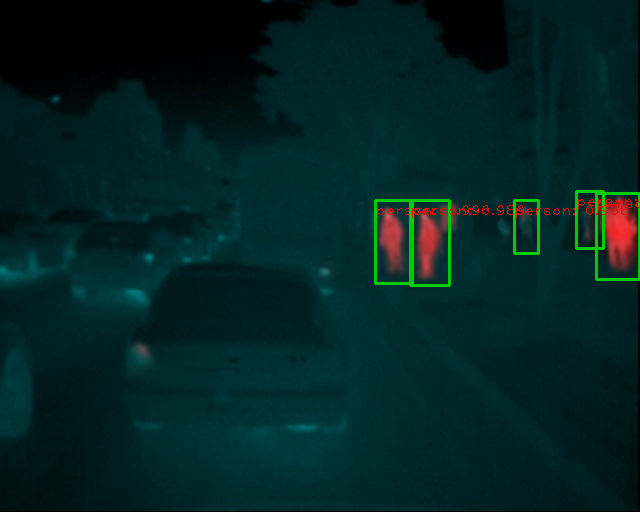} \\
     (f) &
     \includegraphics[width=0.18\linewidth]{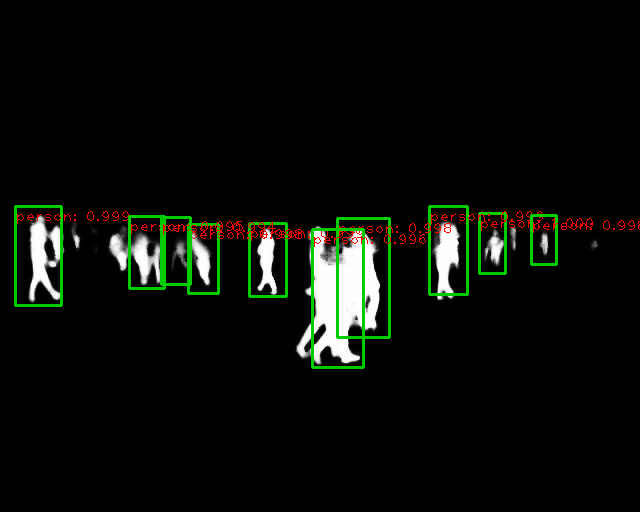} &
     \includegraphics[width=0.18\linewidth]{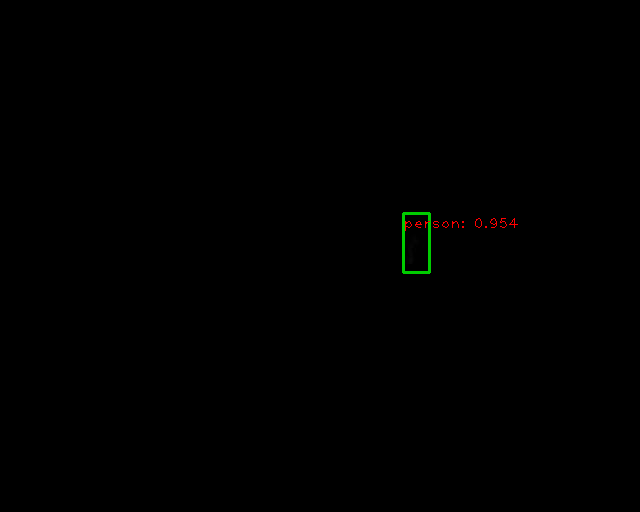} &
     \includegraphics[width=0.18\linewidth]{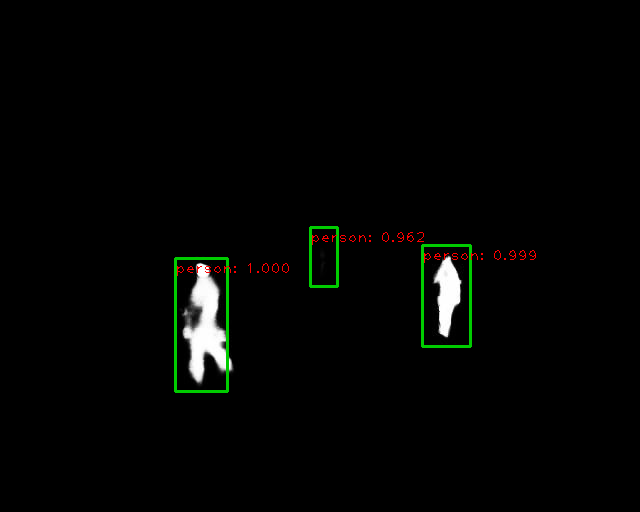} &
     \includegraphics[width=0.18\linewidth]{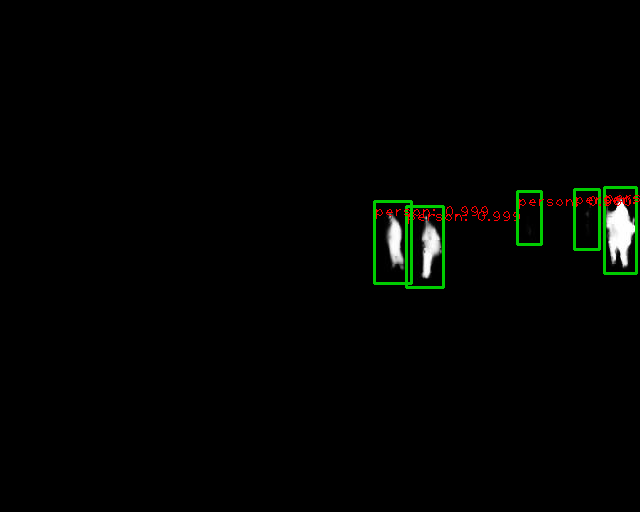} \\
     (g) &
     \includegraphics[width=0.18\linewidth]{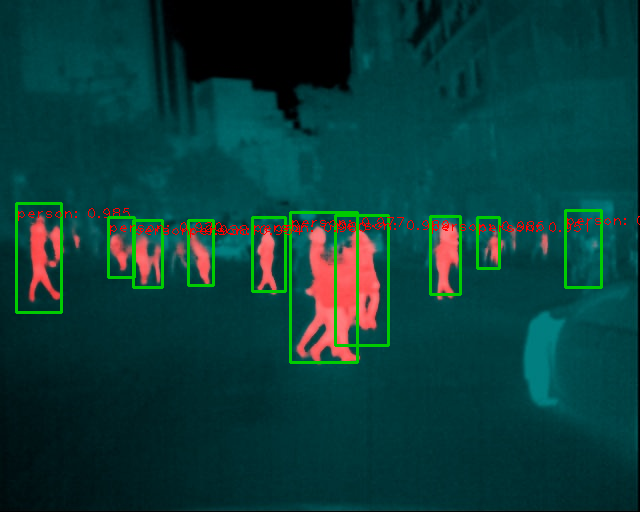} &
     \includegraphics[width=0.18\linewidth]{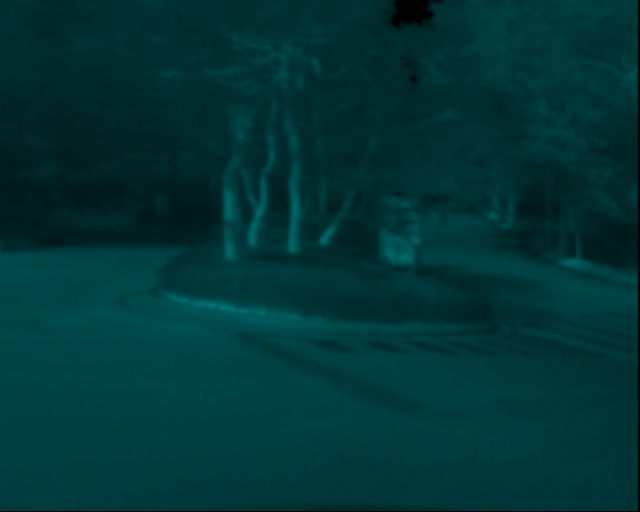} &
     \includegraphics[width=0.18\linewidth]{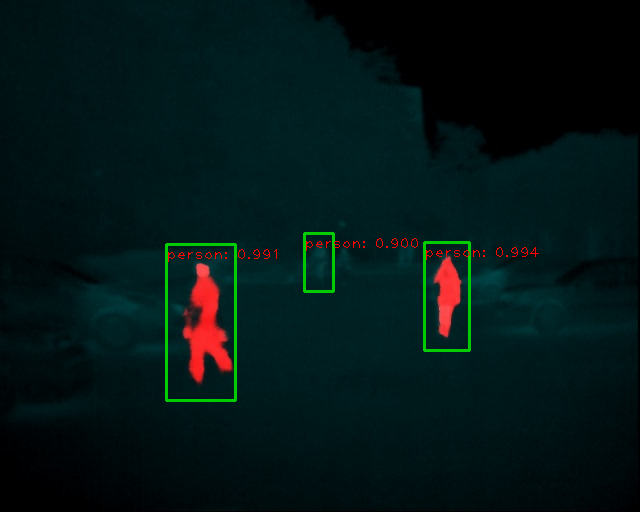} &
     \includegraphics[width=0.18\linewidth]{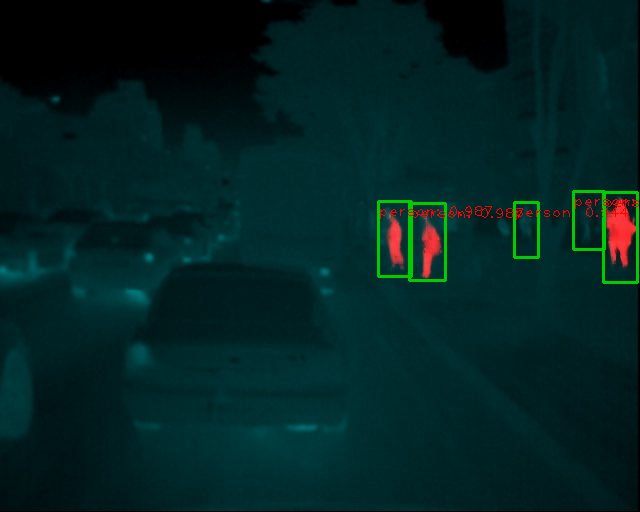} \\
     & 1 & 2 & 3 & 4
\end{tabular}
\caption{\textbf{Sample results from pedestrian detection} on images (1)-(4) from methods: (a) Thermal Images, (b) Static Saliency, (c) Static Saliency + Thermal, (d) PiCA-Net Saliency, (e) PiCA-Net Saliency + Thermal, (f) $R^3$-Net Saliency, (g) $R^3$-Net Saliency + Thermal}
\label{fig:all-results}
\end{figure*}

After evaluating the pedestrian detectors trained separately on thermal images, saliency maps, and thermal images augmented with saliency maps from different techniques, we find that the saliency maps indeed contribute to improved performance. The detector performance for each technique is summarized in Table~\ref{tab:all-results}, and Miss Rate vs FPPI plots are shown in Figure~\ref{fig:fppi_mr}. Below, we discuss some of the important results.
\par\noindent
\textbf{Using only Thermal Images.}
Our baseline detector using only thermal images achieves a miss rate of 44.2\% on the day images and 40.4\% on the night images displaying a large scope for improvement. It is evident from the results however, that thermal images give better performance at nighttime compared to daytime due to low contrast heat maps during the day, as seen in Figure \ref{fig:all-results}(a).
\par\noindent
\textbf{Using Thermal Images with Static Saliency Maps.} The pedestrian detector achieves a miss rate of 39.4\% on day thermal images combined with their static saliency maps, which is an absolute improvement from the baseline by 4.8\%. However, we do not notice any improvement at nighttime, and find this method to have induced a significant number of false positives hurting the precision. This indicates that although static saliency methods show some potential, they are not viable for deployment in round-the-clock applications.
\par\noindent
 \textbf{Using Thermal Images with Saliency Maps generated from Deep Networks.} Our approach augmenting thermal images with deep saliency maps extracted using PiCA-Net achieves a miss rate of 32.2\% for day images and 21.7\% for night images, which is a considerable improvement of 12\% and 18.7\% respectively over the baseline. The approach augmenting saliency maps from $R^3$-Net achieves a miss rate of 30.4\% for day images and 21\% for night images, which is an even better improvement of 13.4\% and 19.4\% over the baseline respectively, as illustrated in Figure~\ref{fig:comparison}. These improvements can be explained by the visualizations in Figure \ref{fig:all-results} which shows that these methods illuminate only pedestrians in the scenes, helping the detector identify pedestrians even under difficult lighting conditions. 
 Moreover, $R^3$-Net achieves a mean Average Precision of 68.5\% during daytime which is a 6.9\% improvement, and 73.2\% during nighttime which is a 7.7\% improvement over the baseline. This suggests that deep saliency methods are useful at all times. 

\subsubsection{Qualitative analysis and effectiveness of saliency maps for Pedestrian Detection}
Figure~\ref{fig:all-results} shows detections on 4 images in different settings using all techniques.
In image 1, we can see that augmenting saliency map 1(b) helps capture the rightmost
missed detection in 1(a), showing its potential in cluttered scenes. In image 2(a), we see a tree detected as a false positive in the thermal image, which is a frequently occurring phenomenon in our observations. Note that the saliency maps in 2(d) \& (f) puts very little emphasis on this region. Therefore, after combining the thermal image with the saliency map, the detector is able to get rid of this false positive (see 2(c), (e) \& (g)). Image (3) shows comparable performance of thermal and saliency detection methods at nighttime. Note that the center-right detection missed in the saliency map in 3(d) was captured in 3(e) after including the thermal information.  In Image 4, the car tail-light captured by the saliency map in 4(d) is removed with the help of information from the thermal image in 4(e); whereas the detection in the middle missed by 4(a) is captured in the deep saliency maps in 4(d) \& (f) and therefore included in the final detections in 4(e) \& (g). This emphasizes the complementary nature of the two techniques, thus confirming our hypothesis. 

\section{Conclusion and Future Work}
\label{sec:conc}
We make two important contributions in this paper. First, we provide pixel level annotations of pedestrian instances on a subset of the KAIST Multispectral Pedestrian dataset. Second, we show that deep saliency networks trained on this dataset can be used to extract saliency maps from thermal images, which when augmented with thermal images, provide complementary information to the pedestrian detector resulting in a significant improvement in performance over the baseline approach.

In this paper, we augmented thermal images with their saliency maps through a channel replacement strategy prior to feeding them into the network. It would be interesting to see if infusing the saliency map into shared layers in the network using a saliency proposal stage, and then jointly learning the pedestrian detection and the saliency detection task similar to SDS R-CNN\cite{SDS-RCNN} would improve the detector performance.
Deep saliency techniques would also benefit from the presence of large amounts of pixel level annotations, indicating a necessary expansion of our dataset. Moreover, saliency techniques used for thermal images are also expected to work for color images and our annotations can be used for the same purpose.


\section*{Acknowledgements}
We would like to thank our peers who helped us improve our paper with their valuable inputs and feedback, in no particular order - Huaizu Jiang, Takeshi Takahashi, Sarim Ahmed, Sreenivas Venkobarao, Elita Lobo, Bhanu Pratap Singh, Joie Wu, Yi Fung, Ziqiang Guan, Aruni Roy Chowdhury and Akanksha Atrey.

{\small
\bibliographystyle{ieee}

}

\end{document}